\documentclass[letterpaper, 10 pt, conference]{ieeeconf}

\IEEEoverridecommandlockouts                              

\usepackage{graphics} 
\usepackage{amsmath,amssymb,amsfonts} 
\usepackage{bm}
\usepackage{subcaption}	 
\usepackage{graphicx}
\usepackage{multirow}
\usepackage{makecell}
\usepackage{array}
\usepackage{caption}
\usepackage{tabularx}
\usepackage{svg}
\usepackage{url}
\usepackage{hyperref}
\hypersetup{colorlinks,urlcolor=red}
\usepackage{xcolor}
\usepackage{threeparttable}
\usepackage{colortbl}
\usepackage[T1]{fontenc}
\usepackage{makecell}  % 支持 \makecell 命令
\usepackage{arydshln}  % 支持 \hdashline 命令

\usepackage{booktabs}
\usepackage{pifont}

\usepackage[backend=bibtex,natbib=true,style=numeric-comp,sorting=none,giveninits=true,maxbibnames=99,url=false,doi=false]{biblatex}
\title{\LARGE \bf
L2Calib: $SE(3)$-Manifold Reinforcement Learning for Robust \\
Extrinsic Calibration with Degenerate Motion Resilience
}

\author{Baorun Li$^{1}$, Chengrui Zhu$^{1}$, Siyi Du$^{1}$, Bingran Chen$^{1}$, Jie Ren$^{1}$, Wenfei Wang$^{3}$, Yong Liu$^{1,2}$, Jiajun Lv$^{1,\dag}$ % <-this % stops a space
\thanks{$^{1}$Institute of Cyber-Systems and Control, Zhejiang University, China.}%
\thanks{$^{2}$State Key Laboratory of Industrial Control Technology, Zhejiang University, China.}
\thanks{$^{3}$Zhejiang Guozi Robotics Technology Co., Ltd.}
\thanks{ $^\dag$Corresponding authors.}
}

\begin{document}

\maketitle
\thispagestyle{empty}
\pagestyle{empty}

\begin{abstract}
Extrinsic calibration is essential for multi-sensor fusion, existing methods rely on structured targets or fully-excited data, limiting real-world applicability. Online calibration further suffers from weak excitation, leading to unreliable estimates. To address these limitations, we propose a reinforcement learning (RL)-based extrinsic calibration framework that formulates extrinsic calibration as a decision-making problem, directly optimizes $SE(3)$ extrinsics to enhance odometry accuracy. Our approach leverages a probabilistic Bingham distribution to model 3D rotations, ensuring stable optimization while inherently retaining quaternion symmetry. A trajectory alignment reward mechanism enables robust calibration without structured targets by quantitatively evaluating estimated tightly-coupled trajectory against a reference trajectory. Additionally, an automated data selection module filters uninformative samples, significantly improving efficiency and scalability for large-scale datasets.
Extensive experiments on UAVs, UGVs, and handheld platforms demonstrate that our method outperforms traditional optimization-based approaches, achieving high-precision calibration even under weak excitation conditions. Our framework simplifies deployment on diverse robotic platforms by eliminating the need for high-quality initial extrinsics and enabling calibration from routine operating data. The code is available at \url{https://github.com/APRIL-ZJU/learn-to-calibrate}.
\end{abstract}

%%%%%%%%%%%%%%%%%%%%%%%%%%%%%%%%%%%%%%%%%%%%%%%%%%%%%%%%%%%%%%%%%%%%%%%%%%%%%%%%
\section{INTRODUCTION}

Extrinsic calibration is critical for multi-sensor systems, ensuring accurate data fusion and spatial alignment. 
Misaligned sensors degrade perception reliability, affecting tasks like object detection, navigation, and simultaneous localization and mapping (SLAM).
Moreover, external factors like mechanical vibrations and environmental shifts necessitate periodic recalibration to maintain long-term operational stability.
% 现有的方法已经取得了一些进展。
Despite significant advancements in sensor-specific calibration techniques such as LiDAR-IMU \cite{lv2022observability} and camera-LiDAR \cite{zhou2018automatic} systems, extrinsic calibration remains a persistent challenge.
Existing offline extrinsic calibration methods typically rely on fully excited sequences \cite{rehder2016general}, stable feature tracking \cite{furgale2012continuous}, and carefully designed calibration targets \cite{fu2019lidar} or structured environments \cite{lv2020targetless}, making the calibration labor-intensive and impractical for non-expert users or real-world deployments. Although accurate, these approaches fail to adapt to dynamic sensor shifts caused by mechanical stress or environmental disturbances, necessitating online calibration.
However, online calibration methods struggle with insufficient motion excitation during routine operations, leading to unreliable calibration. Consequently, developing calibration methodology that ensures stable multi-sensor fusion for navigation and localization holds significant practical importance. 

There are three key insights for a user-friendly calibration method. First, modern robotic platforms integrate diverse sensor configurations, demanding calibration frameworks with cross-modal adaptability. Second, robotic systems generate extensive operational data spanning varied scenarios and motion dynamics, yet existing methods fail to fully exploit them for calibration.
Third, physical constraints often limit motion excitation when collecting data, necessitating robust calibration under suboptimal conditions. 
Therefore, a general, easy-to-use, and robust calibration framework remains elusive.

Reinforcement learning has been extensively applied to exploration in continuous spaces and robotic control problems, demonstrating its capacity for autonomous acquisition of complex skills. Notably, recent advancements have validated its effectiveness in hyperparameter optimization tasks\cite{messikommer2024reinforcement}, showing both computational efficiency and robust performance in high-dimensional parameter spaces. 
However, its potential for extrinsic calibration remains largely unexplored.
% To bridge this gap, we propose a novel reinforcement learning-based calibration framework characterized by exceptional generality and robustness.  Our framework eliminates dependencies on structured environments, specialized calibration targets, or meticulously designed motion sequences, offering ease of deployment across diverse odometry-capable sensor configurations. Key advantages include: $1)$ Effective utilization of readily available reference trajectories $2)$ Full exploitation of operational data collected during routine robotic tasks $3)$ Reliable calibration results even under weak or insufficient excitation conditions. These capabilities prove particularly advantageous in scenarios where acquiring high-quality calibration data is challenging, such as heavy-duty vehicle operations or dynamic unstructured environments.
To bridge this gap, we propose an RL-based extrinsic calibration framework that directly optimizes the $SE(3)$ extrinsics to enhance odometry accuracy. Unlike traditional methods that depend on structured environments or predefined motion sequences, our approach introduces a trajectory alignment reward mechanism, allowing the system to refine extrinsic parameters based on real-world operational data. By modeling 3D rotations using a Bingham distribution, we ensure stable optimization while preserving quaternion symmetry, addressing the inherent challenges of representing rotations in learning-based frameworks. Additionally, our automated data selection module enhances efficiency by filtering uninformative samples, significantly improving scalability across diverse robotic platforms, fully exploring the operational data collected during routine robotic tasks.

The core contributions of our work are as follows:
\begin{itemize}
\item We present a novel RL-based, general extrinsic calibration framework that reformulates 3D rotation representation using Bingham distribution, providing a probabilistic model specifically tailored for the $SE(3)$ manifold, enabling robust joint estimation of rotational and translational parameters. 
\item We propose a trajectory alignment reward mechanism that quantitatively aligns reference trajectories with tightly coupled proprioceptive-exteroceptive odometry estimates, eliminating dependencies on structured environments or specialized calibration targets while maintaining compatibility with arbitrary odometry-capable sensor suites.
\item The framework is capable of joint calibration across multi-sequence datasets to reduce data collection efforts and relax motion excitation requirements, enabling reliable parameter estimation even under insufficient excitation. The proposed automated data selection module filters uninformative samples, significantly improving computational efficiency.
\item  Extensive experiment across handheld, UAV, and UGV platforms demonstrates that the proposed framework outperforms traditional optimization-based methods. It achieves superior calibration accuracy without relying on high-quality initial extrinsic parameters. The method simplifies the calibration process and broadens its applicability. Currently, the framework is adapted for LiDAR-IMU calibration, and we have open-sourced it as a practical tool to encourage further research into automated calibration methods.

\end{itemize}

\section{RELATED WORK}
% 传感器标定领域已经做出了许多awesome的工作。传统的标定方法通常依赖于几何优化、传感器运动先验或已知的环境特征，以估计传感器间的外参。例如Gentil等，使用一组墙面为目标，通过对插值后的IMU测量进行预积分，来表征Lidar的运动失真，以恢复Lidar-IMU的外参。Fu等人，利用环绕摆放的棋盘格或者多边形板完成Lidar和camera的数据关联，并且消除了时间同步的要求。Furgale等人，以棋盘格为标定目标，基于连续时间轨迹理论，提出了一套新的标定框架，完成了时空外参的标定，进一步提升了精度。

% 相比与上述方法，target-free的方法减少了人工的需要，是近年来传感器标定领域的研究热潮。LV等人LI-Calib受连续时间轨迹的启发，开发了一套targetless的lidar-imu标定方法；其后添加了能观性的分析，共同的优化内参和时空外参；Chen等人将其连续时间的方法应用到更多传感器上，并且推广到lidar-camera-imu-gnss的情况下。这些方法需要首先依据IMU raw measurement然而，这些基于优化的方法依赖于对外参有一个较好的猜测以作为初值。

% Zhu等人LI-init，利用两传感器之间角速度、加速度关系对齐对外参进行优化，其lidar角速度和加速度测量由匀速假设差分得到，由于标定数据录制时需要充分的激励，因此通常无法保证这个假设总是成立；更多的，基于外参的准确性影响里程计的表现这个前提，许多工作将外参纳入到EKF状态量中，与位姿和传感器测量共同估计。然而这种方法依赖于里程计前端特征提取的鲁棒性，里程计前端并不是一个能作为真值去eval外参好坏的metric。

Sensor calibration has been a focal point in robotics research, with numerous significant contributions addressing the estimation of extrinsic parameters between heterogeneous sensors. Traditional calibration methods typically rely on geometric optimization, prior knowledge of sensor motion, or known environmental features to estimate the extrinsic parameters. For example, Le et al.\cite{le20183d} proposed a method that uses a set of walls as calibration targets, using preintegrated IMU measurements to characterize LiDAR motion distortion and recover extrinsic parameters between LiDAR and IMU. Similarly, Fu et al.\cite{fu2019lidar} utilized surrounding checkerboards or polygonal boards to establish data association between LiDAR and cameras, eliminating the need for strict temporal synchronization. Furgale et al.\cite{furgale2012continuous,furgale2013unified} introduced a novel calibration framework based on continuous-time batch optimization, using checkerboards as calibration targets to achieve spatiotemporal extrinsic calibration, further enhancing calibration accuracy. Rehder et al.\cite{rehder2014spatio,rehder2016general} took a similar approach to calibrating multi-sensor systems. Although this approach significantly improves the accuracy of the calibration, it still relies heavily on carefully designed calibration targets and assumes a controlled environment, which may not always be feasible in real-world applications. These limitations highlight the need for more flexible and target-free methods.

In contrast to traditional approaches, recent research has turned towards target-free calibration methods, which significantly reduce the reliance on artificial calibration targets. This shift simplifies the calibration process and enhances the system's flexibility. Lv et al.\cite{lv2020targetless} proposed LI-Calib, a targetless LiDAR-IMU calibration method inspired by continuous-time trajectory theory. This approach was later extended to incorporate observability analysis and the calibration of intrinsic parameters in \cite{lv2022observability}. Chen et al.\cite{10288412,10848334}further generalized the continuous-time method to multi-sensor systems, including LiDAR, camera, IMU, and GNSS. However, these optimization-based methods rely on IMU raw measurements to recover the initial motion trajectory, which can be biased due to IMU measurement noise. Additionally, they require a reasonably accurate initial guess for the extrinsic parameters to avoid convergence to local optima, posing challenges in practical applications.
Recent advances have explored alternative strategies to address these limitations. Zhu et al.\cite{zhu2022robust} proposed LI-init, which optimizes extrinsic parameters by aligning angular velocity and acceleration measurements between LiDAR and IMU. But it is still limited by the constant velocity assumption, which may not hold during calibration data collection.

\section{METHODOLOGY}
\begin{figure*}
    \centering
    \includegraphics[width=1\linewidth]{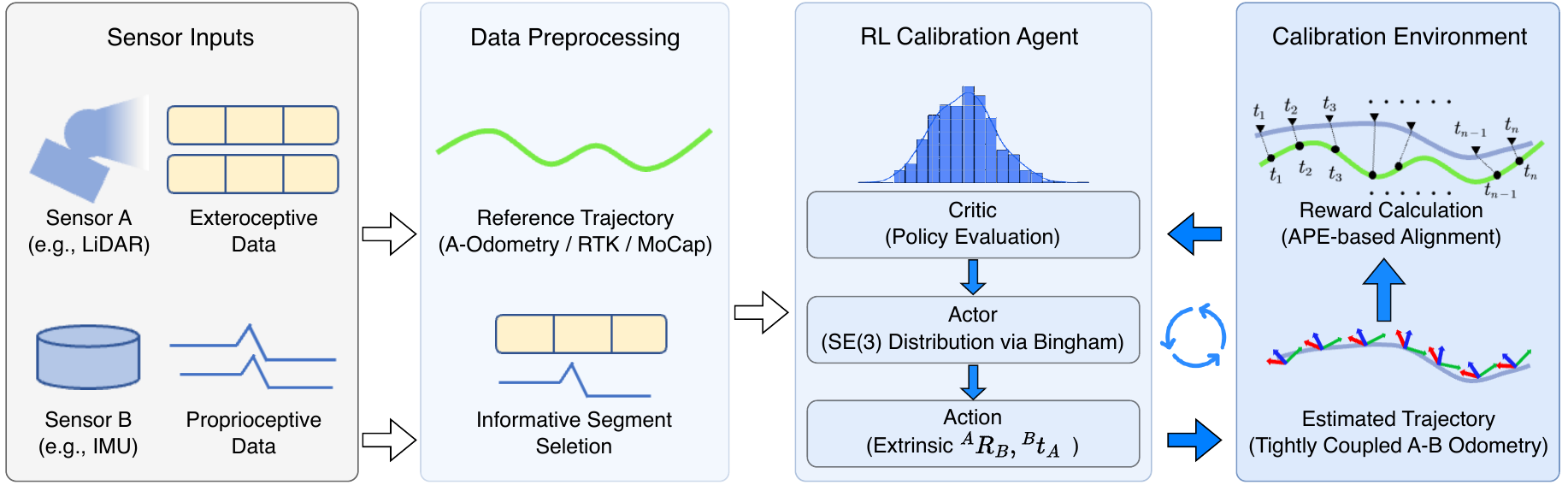}
    \caption{Reinforcement Learning-Based Sensor Calibration Framework. 
    %The system comprises threecomponents: Data Preprocessing, Calibration Agent, and Calibration Environment. Given data from an exteroceptive sensor (e.g., LiDAR) and a proprioceptive sensor (e.g., an IMU), the system first constructs a reference trajectory from either the exteroceptive sensor based odometry or ground truth (e.g., GPS or MoCap). Informative segments containing synchronized data from both sensors are selected by the Data Preprocessing module and passed to a tightly-coupled odometry in the Calibration Environment to estimate trajectories. The goal of the RL agent, designed with an Actor-Critic structure, is to find an optimal extrinsic parameter on the $SE(3)$ manifold that minimizes APE error, comparing the estimated trajectory against the reference trajectory.
    An Actor-Critic agent iteratively refines extrinsic parameters on $SE(3)$ by minimizing trajectory error between tightly-coupled LiDAR-IMU odometry and a LiDAR-based reference across informative data segments.
    }
    \label{fig:pipeline}
    \vspace{-1em}
\end{figure*}

The pipeline of the proposed $SE(3)$-Manifold RL-based extrinsic calibration method is illustrated in Fig. \ref{fig:pipeline}. 
First, the calibration sequence is highly flexible, supporting both specially recorded sequences and data collected during routine operations. The \textit{reference trajectories}, essential for reward computation, can be obtained from diverse sources including exteroceptive sensor-based odometry (e.g., Visual Odometry or LiDAR Odometry), Real-Time Kinematic (RTK) positioning systems, or motion capture systems. The proposed RL-based calibration method then uses iterative agent-environment interactions to learn the extrinsics. At each iteration, the agent samples rotation parameters from the Bingham distribution and translation parameters from a Gaussian distribution (Sec. \ref{sec:policy_param}). The agent applies the samples to compute the \textit{estimated trajectory} from a tightly-coupled odometry system (e.g., LIO, VIO). After aligning the reference and estimated trajectories, the Absolute Pose Error (APE) is calculated as the reward to guide the optimization of extrinsics (Sec. \ref{sec:reward}). Additionally, an optional data selection module automatically filters informative data segments based on motion observability criteria, ensuring computational efficiency and robustness in degraded motion scenarios (Sec. \ref{sec:data_selection}).

\subsection{Problem Formulation}\label{ProblemFormulation}

Extrinsic calibration in non-degenerate scenarios has been extensively studied; however, research on calibration under degenerate conditions remains limited. Existing approaches attempt to address such cases with heuristic constraints. For instance, GRIL-Calib~\cite{10506583} incorporates ground constraints for LiDAR-IMU calibration but is restricted to ground vehicle applications. Similarly, OA-LICalib~\cite{lv2022observability} truncates gradients along degenerate axes, which does not provide an optimal solution. We propose a paradigm shift by reformulating extrinsic calibration as a decision-making problem under a reinforcement learning framework, where degeneracy-aware exploration naturally emerges from the learning process. Considering that robotic systems often integrate GPS receivers or operate in motion-capture-equipped environments to acquire ground-truth trajectories. In the absence of absolute references, exteroceptive sensor-based odometry can provide reliable trajectory estimates.
\begin{figure}
    \centering
    \includegraphics[width=1\linewidth]{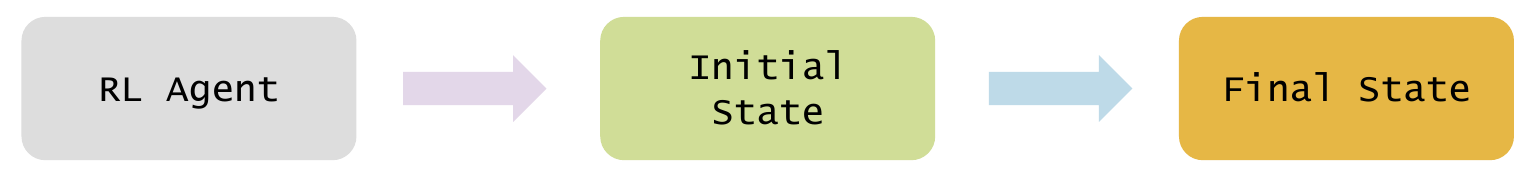}
    \caption{A single interaction between the RL Agent and the Environment. In this interaction, the Agent generates an action, which in this context corresponds to the extrinsic parameters. This action is applied to the initial state, causing it to transition to the final state. The initial state is defined as the state before an odometry update is performed, while the final state is the state after the odometry update.}
    \label{fig:mdp}
    \vspace{-1em}
\end{figure}
Given measurements from an exteroceptive sensor $A$ and an inertial sensor $B$, and a reference trajectory $T_{ref}(t) \in SE(3)$ from the sources listed above, we aim to estimate the extrinsic ${}^AT_B \in SE(3)$ by minimizing the pose error between the transformed sensor trajectory $T_A(t)$ and $T_{ref}(t)$:
\begin{equation}
X^* = \underset{X}{\arg\min} \sum_{t=1}^N \rho \left( T_{ref}(t)^{-1} \cdot T_B(t) \cdot {}^AT_{B}^{-1} \right),
\end{equation}
where $\rho(\cdot)$ denotes a robust cost function. We recast this problem as a \textit{Markov Decision Process} (MDP) $\mathcal{M}=(\mathcal{S,A,R,P})$. Unlike standard MDPs, which involve sequential decision-making, we model the calibration process as a \textit{single-step decision problem}. 
The state space $\mathcal{S}$ consists only of the initial state and the terminal state, as shown in Fig. \ref{fig:mdp}. while the action space $\mathcal{A}$ constitutes the $SE(3)$ manifold, with each action corresponding to a candidate extrinsic. The reward function $\mathcal{R}$ is designed to be negative correlation to the pose error, providing a quantitative metric for evaluating the effectiveness of the current action. The transition function $\mathcal{P}$ is deterministic, mapping all states with probability 1 to the terminal state upon action execution.

\subsection{Policy Parameterization}
\label{sec:policy_param}

\begin{figure}[t]
    \centering
    \subfloat[Gaussian Sampling]{%
        \includegraphics[width=0.6\linewidth]{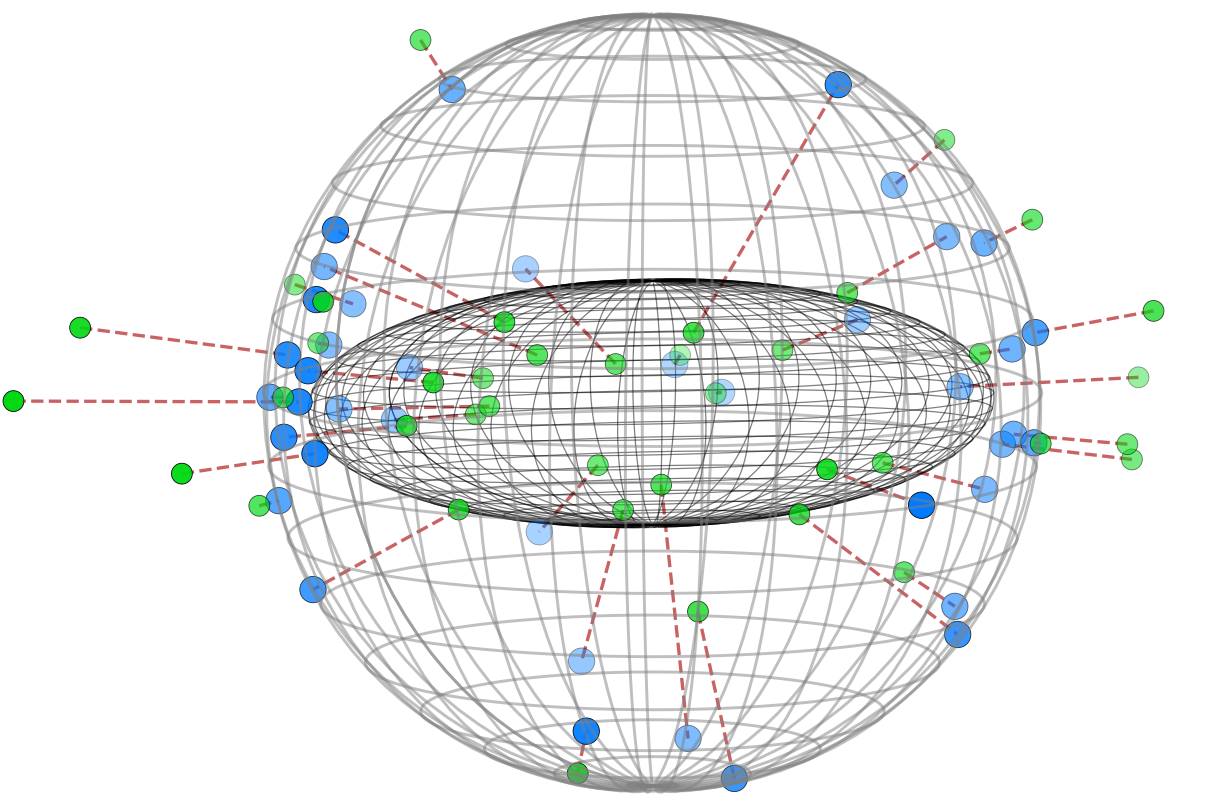}%
    }
    \subfloat[Bingham Sampling]{
        \includegraphics[width=0.39\linewidth]{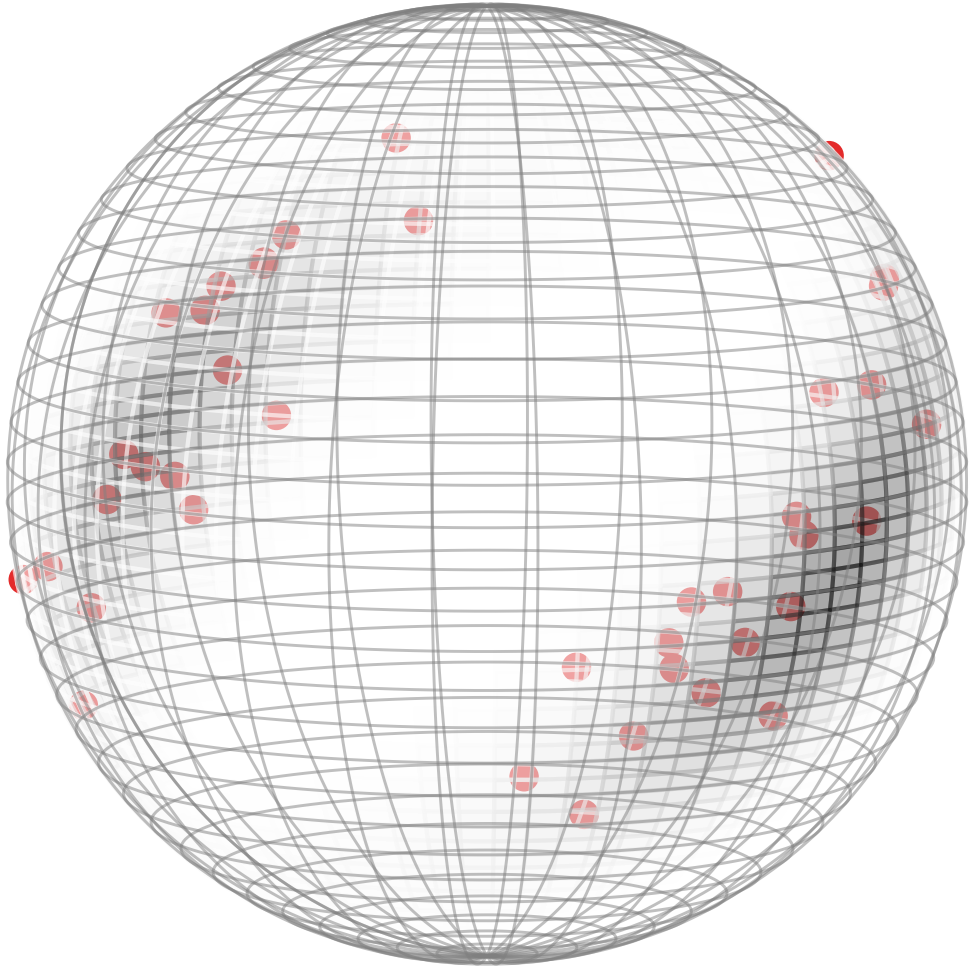}
    }
    \caption{Rotation sampling comparison. (a) Gaussian-based rotation sampling: The ellipsoid represents the 1-$\sigma$ contour of a Gaussian distribution in $\mathbb{R}^3$. Samples (green) are drawn from a 3D Gaussian, projected onto the unit sphere (blue) along radial directions (red dashed). This indirect process distorts the target distribution due to non-uniform projection. (b) Bingham-based rotation sampling: Samples (red) are directly from a Bingham distribution defined on the spherical manifold, with shaded regions indicating the probability density with concentration parameters $\lambda=(-30,-4,0)$. The distribution is inherently antipodally symmetry, where any sample and its diametrically opposite point have equal probability.}
    \label{fig:distribution_visualization}
    \vspace{-1em}
\end{figure}

The action space encompasses both rotational and translational extrinsic parameters, constituting a representation on the $SE(3)$ manifold. In practice, this is expressed as the Cartesian product of quaternions on $S^3$ and vectors in $\mathbb{R}^3$. Conventional reinforcement learning methods typically use Gaussian distributions to parameterize each component of the extrinsics. However, sampling from a Gaussian distribution in $\mathbb{R}^4$ yields a four-dimensional vector, not a quaternion. This vector must then be projected onto the unit hypersphere to obtain a valid quaternion. However, the quaternions produced by this process may not effectively be drawn from a single, consistent probability distribution, so directly sampling rotation parameters from a Gaussian distribution in $\mathbb{R}^4$ is not an optimal representation for quaternions, which neglects the inherent bi-modality and symmetry of quaternions\cite{james2022bingham}. For a more intuitive, albeit less rigorous, explanation, Fig. \ref{fig:distribution_visualization} provides a visualization on the 3D space that illustrates why the Bingham distribution is more suitable for representing distributions of quaternions. Inspired by \cite{james2022bingham}, we model the probability distribution of the rotational component directly on the $S^3$ manifold using the Bingham distribution \cite{10.1214/aos/1176342874}, which provides a more suitable representation, leading to faster and more stable convergence during optimization.

Specifically, the probability density function (PDF) of the Bingham distribution in $S^3$ is given by
\begin{equation}
    p(\mathbf{x};\mathbf{M},\mathbf{Z})=\frac{1}{N(\mathbf{Z})}\exp\left(\mathbf{x}^\mathsf{T}\mathbf{M}\mathbf{Z}\mathbf{M}^\mathsf{T}\mathbf{x}\right),
\end{equation}
where $\mathbf{x}\in S^3$ denotes a unit quaternion, $\mathbf{M}$ is an orthogonal matrix, and $\mathbf{Z} \in \mathbb{R}^{4\times4}$ is a diagonal matrix representing the dispersion of the distribution, given by $diag(z_1, z_2,z_3,0)$, $z_1\leq z_2\leq z_3\leq 0$. $N(\mathbf{Z})$ is the normalization constant. Notably, the probability density function exhibits antipodal symmetry: $p(\mathbf{x})=p(-\mathbf{x})$ for all $\mathbf{x}\in S^3$, this is consistent with the behavior of quaternions. Moreover, if $u$ is a mode of the distribution, then $-u$ is also a mode \cite{glover2014quaternion}. These intrinsic symmetries ensure the Bingham distribution properly accommodates the topological structure of the $S^3$ manifold, particularly its antipodal identification requirement in rotation representation.

To implement rotational policy parameterization via Bingham distributions, efficient sampling from this antipodally symmetric distribution becomes essential. We employ the acceptance-rejection sampling framework introduced in \cite{kent2013new}, utilizing an Angular Central Gaussian (ACG) distribution as the proposal density. The ACG density function on the $q-1$ dimensional hypersphere $S^{q-1}$ is formally expressed as:
\begin{equation}
p(\mathbf{x},\mathbf{\Lambda})=\sqrt{|\mathbf{\Lambda}|}(x^\mathsf{T}\mathbf{\Lambda} x)^{-\frac{q}{2}}, 
\end{equation}
where $\mathbf{\Lambda} \in \mathbb{R}^{q\times q}$ denotes a symmetric positive-definite concentration matrix. Optimal envelope matching between the ACG and Bingham distributions is achieved when the concentration matrices satisfy the parameter coupling relationship:
\begin{equation}
% \begin{aligned}
    \mathbf{\Lambda}=\mathbf{I}+\frac{2\mathbf{M}\mathbf{Z}\mathbf{M}^T}{b} \,,
    % s.t. \sum_i^n\frac{1}{b+2z_i}=1.
% \end{aligned}
\end{equation}
where $b$ is an integer satisfying $\sum_i^n\frac{1}{b+2z_i}=1$.

\subsection{Reward Computation}\label{sec:reward}
We employ Umeyama alignment \cite{umeyama1991least} to align the reference trajectory with the estimated trajectory. Let \(^rT_A(t)\) denote the aligned reference trajectory and \(^eT_B(t)\) the estimated trajectory. The reward \(\mathcal{R}\) is computed as follows
\begin{equation}
\begin{aligned}
    e_t(i) &= |^eR_{B}(i){}^Bt_A + {}^et_B(i) - {}^rt_A(i)|, \\
    e_r(i) &= \cos^{-1}\left(\frac{\text{tr}\left({}^eR_B(i){}^BR_{A}{}^rR_A^{T}(i)\right) - 1}{2}\right), \\
    \mathcal{R} &= \exp\left(-\sqrt{\sum_i \frac{e_t(i)}{4N}} - \sqrt{\sum_i \frac{e_r(i)}{\pi^2 N}}\right),
\end{aligned}
\end{equation}
where $N$ denotes the number of corresponding poses. The \(^rR_A(i)\) and \(^eR_B(i)\) represent the rotational components of the reference and estimated poses at time \(i\), respectively, while \(^rt_A(i)\) and \(^et_B(i)\) denote the translational components. The translational and rotational pose errors are quantified by \(e_t(i)\) and \(e_r(i)\), respectively. Since translations and rotations have different magnitudes, we apply normalization to mitigate the impact of this asymmetry on the computed error.

% Note that the reference trajectory can be obtained from various sources, such as LiDAR odometry (LO), visual odometry (VO), RTK/GNSS systems, or motion capture systems. 

\subsection{Data Selection}
\label{sec:data_selection}

\begin{figure}
\centering
\includegraphics[width=1\linewidth]{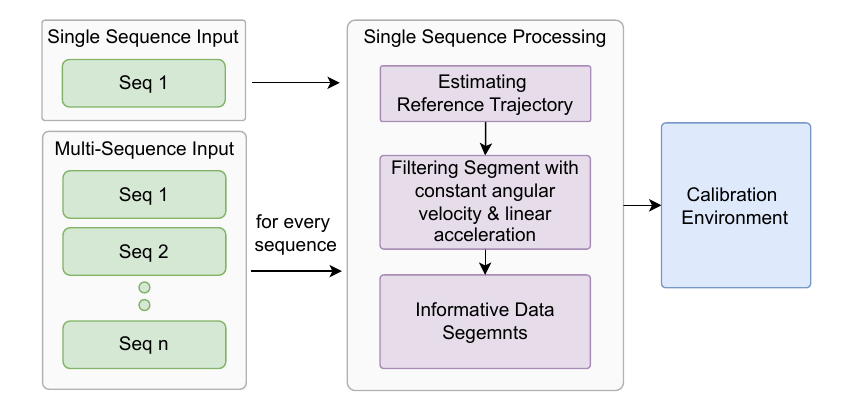}
\caption{Data selection module for single-sequence and multi-sequence inputs, which filters raw sensor data and extracts informative segments.}
\label{fig:data selection}
\vspace{-1em}
\end{figure}

% A common challenge in extrinsic calibration is the need for carefully designed motion sequence with sufficient excitation, which often requires significant expertise and effort to record. To tackle this issue, our calibration framework is designed to handle both single-sequence-input(for which the data is fully excited) or multi-sequence-input(long sequences of data collected during the robot's routine operations or freely recorded in real-world scenarios).
% For the multi-sequence input scenario, a data selection module is essential to eliminate unnecessary computations. Ideally, an observability criterion should be employed to determine which data are informative. However, this approach still requires computing the derivatives of each odometry estimation, as discussed in \ref{ProblemFormulation}. In practice, we find that simply ensuring non-constant angular velocity and linear acceleration is sufficient to achieve effective selection \cite{yang2020online}.
% Specifically, we first fit a continuous trajectory using the method proposed by \cite{lv2023continuous}, enabling the extraction of angular velocity and linear acceleration as described in \cite{sommer2020efficient}.Then we construct a Jacobi matrix in the following form,

A critical challenge in extrinsic calibration is collecting sequences with sufficient excitation. This is often difficult and introduces additional issues, such as limited movements of heavy vehicle platforms, excessive motion leading to image blurring, or LiDAR point cloud distortion that affects data association. Our method naturally supports both single-sequence and multi-sequence inputs, as illustrated in Fig. \ref{fig:data selection}, enabling calibration across sequences from different sources and reducing data dependency.

In the multi-sequence case, the data selection module plays a crucial role in eliminating unnecessary computations. We adopt a simple yet effective selection criterion: ensuring that the data includes non-constant angular velocity and linear acceleration, thereby ensuring the informativeness of the data \cite{yang2020online}. Given IMU measurements, we construct two Jacobi matrices as follows
\begin{equation}
\mathbf{J}_r=\left[ \begin{array}{c}
    \vdots\\
    \boldsymbol{\omega }_{I_k}\\
    \vdots\\
\end{array} \right] ,\mathbf{J}_t\quad =\left[ \begin{array}{c}
    \vdots\\
    \Omega _{I_k}\\
    \vdots\\
\end{array} \right] ,
\end{equation}
where $\boldsymbol{\omega }_{S_k}$ and $\Omega _{S_k}$ denote the $k$th angular velocity and linear acceleration of the imu frame. By computing the minimum eigenvalues of $\mathbf{J}_r^{T}\mathbf{J}_r$ and $\mathbf{J}_t^{T}\mathbf{J}_t$, we can assess whether a segment contains sufficient information and filter out uninformative segments.

\section{EXPERIMENT}
The proposed RL-based extrinsic calibration method, termed as L2Calib (short for Learn To Calibrate), is applicable to various inertial-based systems, including visual-inertial, LiDAR-IMU and LiDAR-camera-IMU configurations. It introduces a trajectory alignment reward formulation that is compatible with any sensor configuration capable of odometry estimation, ensuring broad applicability. To simplify the experiments and facilitate quantitative evaluation, we focus on the LiDAR-IMU system for validation and compare our approach against two state-of-the-art calibration methods, LI-Init \cite{zhu2022robust} and iKalibr \cite{10848334}. LI-Init is the primary benchmark due to its online calibration capability. L2Calib  employs Bingham distribution for rotation policy representation and includes the data selection module, whereas L2Calib-g uses Gaussian distribution for rotation policy representation.

\begin{table}[tb]
    \centering
    \caption{Overview of the datasets.}
    \label{tab:dataset}
    \resizebox{\columnwidth}{!}{
    \begin{tabular}{lccc}
        \toprule
        \textbf{Dataset} & \textbf{Avg. Duration} & \textbf{Avg. Angular Vel} & \textbf{Extrinsics} \\
        & (s) & ($rad\cdot s^{-1}$) & (m) \\
        \midrule
        CSC       & 45  & 2.204  & [0.650, -0.372, -0.016]  \\
        NTU VIRAL~\cite{nguyen2022ntu} & 361 & 0.138  & [0.050, 0.000, -0.055]  \\
        MCD~\cite{mcdviral2024}       & 509 & 0.338  & [-0.049, -0.031, -0.018] \\
        \bottomrule
    \end{tabular}
    }
\end{table}

We first assess calibration accuracy under well-excited motion using three LiDAR-IMU sequences collected with a self-assembled handheld rig as shown in Fig. \ref{fig:setup}. Additionally, we evaluate the methods on two public datasets, NTU VIRAL \cite{nguyen2022ntu} and MCD \cite{mcdviral2024}, which contain UAV and autonomous vehicle sequences recorded during routine operations. These datasets provide diverse motion patterns, including cases with weak excitation. Table \ref{tab:dataset} summarizes key dataset characteristics, including sequence duration, average angular velocity norm, and extrinsic ground truth. 
Furthermore, to analyze the impact of individual components, we conduct ablation studies on key aspects, including rotation parameterization, where we validate the advantages of our Bingham distribution-based rotation policy modeling, data selection, which filters weakly excited segments to improve efficiency, and reference trajectory sources, where odometry and groudtruth trajectory are compared for their influence on calibration accuracy.

In all experiments, we employ Traj-LO \cite{zheng2024traj} for reference trajectory estimation and Fast-lio2 \cite{xu2022fast} for the tightly-coupled odometry, and initialize extrinsics randomly to demonstrate robustness to poor priors. For training the agent, we use the on-policy algorithm Proximal Policy Optimization (PPO) \cite{schulman2017proximal}. 
To evaluate extrinsic calibration accuracy, we compute 6-DoF errors: translation error as the absolute difference between the ground truth and the estimated, and rotation error by converting the error quaternion $q_{err} = q_{gt}^{-1} \otimes q_{est}$ to Euler angles, where $q_{gt}, q_{est}$ are the ground-truth and estimated, respectively. $\otimes$ represents quaternion multiplication.

\subsection{Handheld Dataset}
\begin{figure} 
    \centering
    \includegraphics[width=0.8\linewidth]{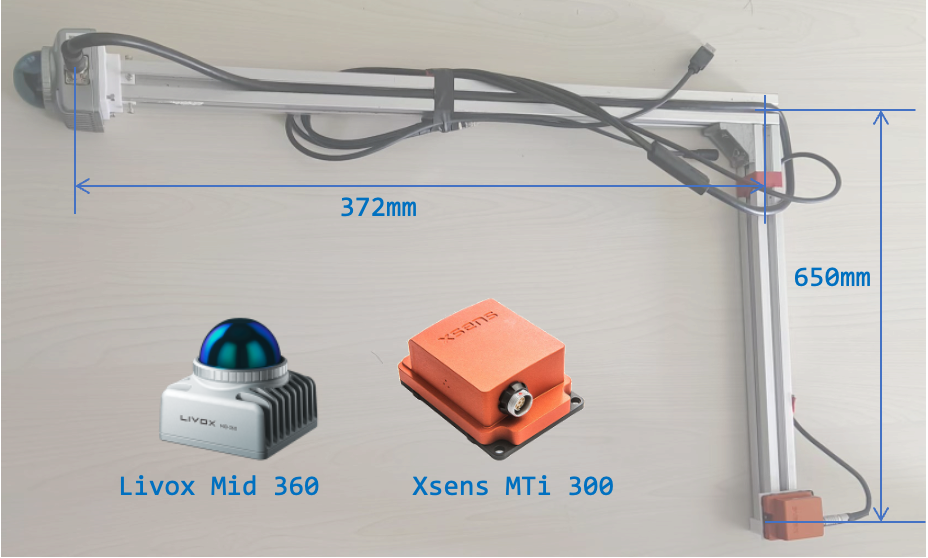}
    \caption{The self-assembled device with the extrinsics listed.}
    \label{fig:setup}
    \vspace{-1em}
\end{figure}

% The ground truth of the extrinsic from LiDAR to IMU is $\left( 0.65025, -0.37250, -0.01600, 180^{\circ}, 90^{\circ}, 0^{\circ} \right)$.
Three fully excited sequences, named \textit{csc\_}0x, are collected in outdoors using our self-assembled handheld rig consisting of a MID-360 LiDAR and a MTi-300 IMU as the physical setup shown in Fig. \ref{fig:setup}. These sequences provide a controlled setting to analyze calibration performance when the sensor motion is sufficiently diverse. The two sensors were deliberately placed considerably apart to assess the methods' robustness and sensitivity to initial extrinsic parameters. 
% The data were collected in an outdoor environment, where the device was handheld, and we deliberately swayed it with an average angular velocity of $2.204$ radians per second, sufficiently exciting the sensor along all 6 axes. We recorded three sequences, named csc\_01, csc\_02, and csc\_03. 

\begin{figure*}[tb]
    \centering
    \begin{subfigure}[b]{0.325\linewidth}
        \centering
        \includegraphics[width=\linewidth]{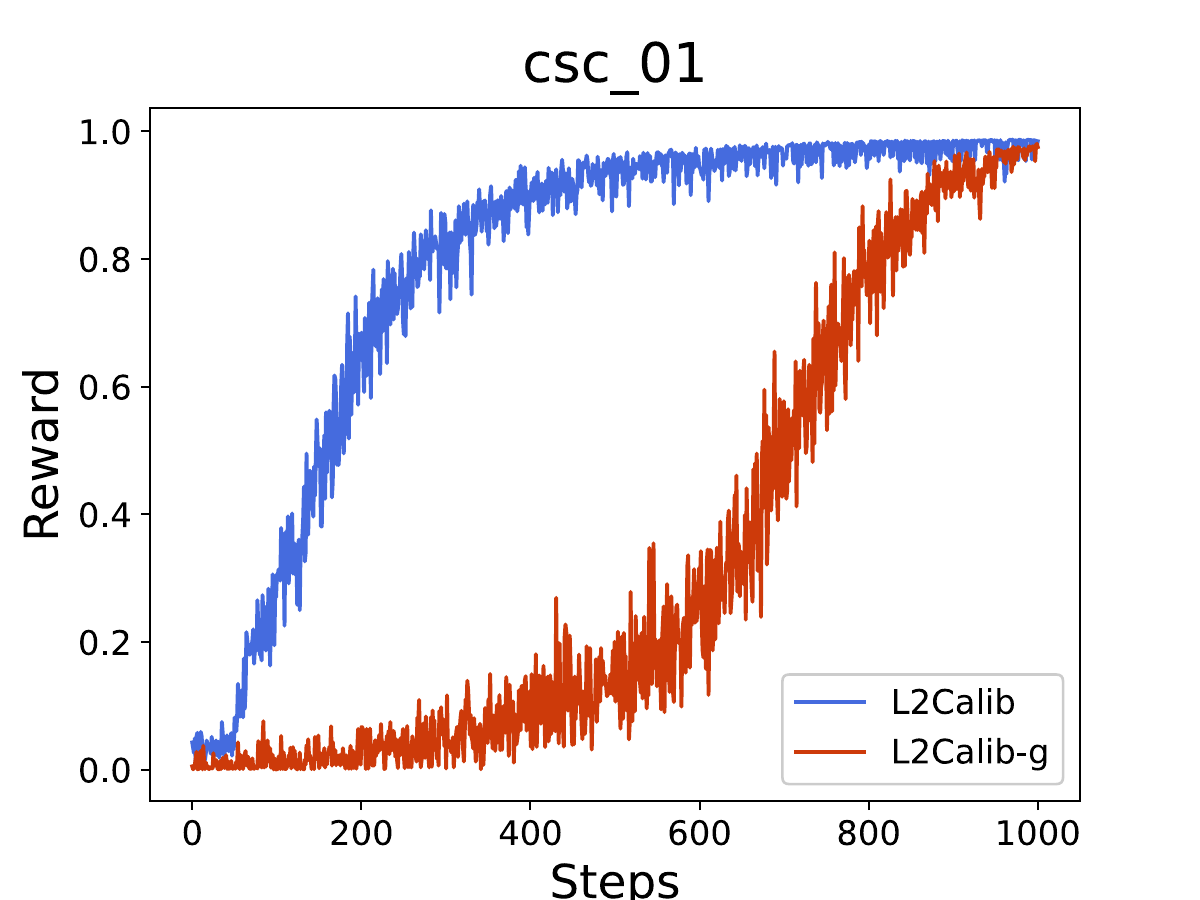}
    \end{subfigure}
    \begin{subfigure}[b]{0.325\linewidth}
        \centering
        \includegraphics[width=\linewidth]{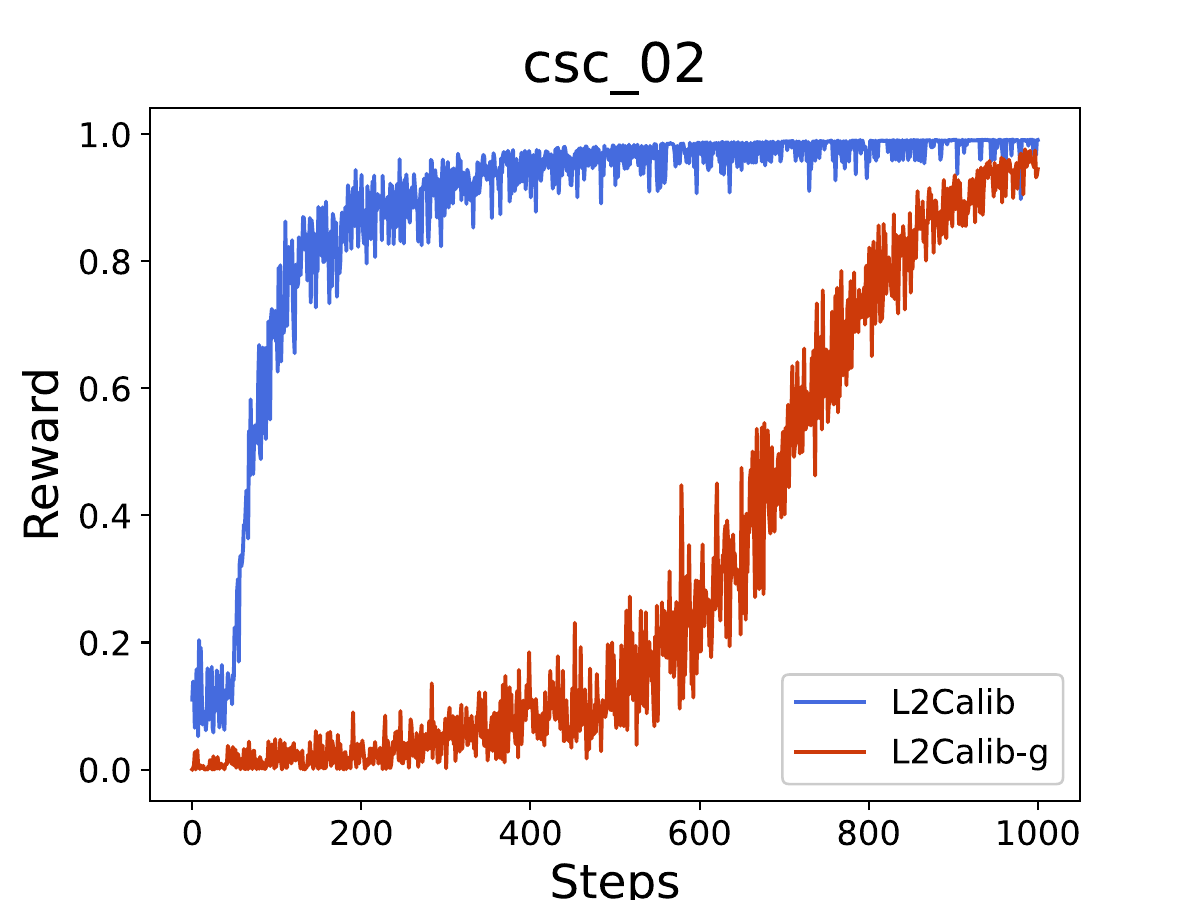}
    \end{subfigure}
    \begin{subfigure}[b]{0.325\linewidth}
        \centering
        \includegraphics[width=\linewidth]{image/exp/fully_excitation/csc_02.pdf}
    \end{subfigure}
\caption{Comparison of convergence speed on fully-excited data using different policy parameterization.}
\label{fig:fully_reward}
\vspace{-2em}
\end{figure*}

Table \ref{tab:handholdcalib} summarizes the calibration results including translation/rotation/average calibration error. L2Calib achieves comparable calibration accuracy compared to existing optimization-based methods.
Fig. \ref{fig:fully_reward} presents the reward evolution curves for rotation policies represented by Gaussian and Bingham distributions. The Bingham distribution-based policy outperforms the Gaussian distribution-based policy in both calibration accuracy and convergence speed, demonstrating the effectiveness of our proposed $SE(3)$-manifold learning method.

% this experiment demonstrates that our proposed RL-based calibration method achieves calibration capabilities in standard scenarios, in the results, there is only one sequence for which the accuracy of the rotational external parameter calibration is not optimal. with accuracy that is either superior or comparable to the state-of-the-art traditional methods.

\begin{table}[tb]
\centering
\caption{Calibration errors of different methods on Handheld sequences. Best results are in bold. L2Calib-g represents using Gaussian parameterization for rotation policy.}
\label{tab:handholdcalib}
\resizebox{\columnwidth}{!}{
\begin{tabular}{@{}c l c c c c c c c c@{}}
\toprule
\multirow{2}{*}{Seq} & \multirow{2}{*}{Method} & \multicolumn{3}{c}{Trans. Error (m)} & \multicolumn{3}{c}{Rot. Error (${}^\circ$)} & \multicolumn{2}{c}{Ave. Error} \\
\cmidrule(lr){3-5} \cmidrule(lr){6-8} \cmidrule(lr){9-10}
                        &                         & x      & y      & z      & roll      & pitch      & yaw      & Trans (m) & Rot (${}^\circ$) \\ \midrule
\multirow{4}{*}{csc\_01} & iKalibr                 & 0.013  & 0.023  & 0.023  & 0.162  & 0.886  & 1.055  & 0.020     & \textbf{0.701}           \\
                        & LI-Init                 & 0.022  & 0.025  & 0.060  & 0.630  & 2.244  & 0.398  & 0.035     & 1.091           \\
                        & L2Calib-g                & 0.028  & 0.025  & 0.018  & 0.905  & 1.109  & 1.046  & 0.024     & 1.020           \\
                        & L2Calib                    & 0.014  & 0.012  & 0.028  & 0.356  & 1.486  & 1.166  & \textbf{0.018}     & 1.003           \\ \midrule
\multirow{4}{*}{csc\_02} & iKalibr                 & 0.020  & 0.021  & 0.026  & 0.134  & 0.986  & 1.074  & 0.022     & 0.731           \\
                        & LI-Init                 & 0.031  & 0.006  & 0.045  & 0.719  & 2.581  & 0.725  & 0.027     & 1.342           \\
                        & L2Calib-g                & 0.029  & 0.016  & 0.025  & 0.333  & 0.951  & 0.012  & 0.023     & 0.432           \\
                        & L2Calib                    & 0.005  & 0.007  & 0.008  & 0.353  & 0.082  & 0.288  & \textbf{0.007} & \textbf{0.241}  \\ \midrule
\multirow{4}{*}{csc\_03} & iKalibr                 & 0.017  & 0.024  & 0.022  & 0.172  & 0.960  & 1.070  & 0.021     & 0.734           \\
                        & LI-Init                 & 0.007  & 0.038  & 0.056  & 0.692  & 1.003  & 1.286  & 0.034     & 0.593           \\
                        & L2Calib-g                & 0.057  & 0.006  & 0.018  & 0.475  & 0.159  & 1.297  & 0.027     & 0.644           \\
                        & L2Calib                    & 0.017  & 0.013  & 0.028  & 0.070  & 0.824  & 0.637  & \textbf{0.020}     & \textbf{0.514}           \\ \bottomrule
\end{tabular}
}
    % \begin{tablenotes}
    %     \footnotesize
    %     \centering 
    %     \item 
    %     The calibration error on each axes and the mean error of translation and rotation. Gaussian represents the Gaussian parameterization for rotation policy in our proposed method. 
    % \end{tablenotes}
\end{table}

% The fully excited experiments demonstrated that our method performs well in standard calibration scenarios, in the results, there is only one sequence for which the accuracy of the rotational external parameter calibration is not optimal. 

%  nine sequences (eee, nya, sbs) from the NTU VIRAL SLAM dataset. Specifically, we calibrated the extrinsic parameters between a horizontal LiDAR (Ouster OS1-16) and an IMU (VectorNav VN100), using the ROS topics /os1\_cloud\_node1/points and /imu/imu, respectively. 

\begin{table}[tb] 
\centering
\caption{Calibration errors on UAV dataset. Best results are in bold. $\times$ indicates the divergent calibration results and ``All Seq'' represents calibration using all sequences. L2Calib-g represents using Gaussian parameterization for rotation policy.}
\label{tab:ntucalib}
\resizebox{\columnwidth}{!}{
\begin{tabular}{@{}c l c c c c c c c c@{}}
\toprule
\multirow{2}{*}{Seq} & \multirow{2}{*}{Method} & \multicolumn{3}{c}{Trans. Error (m)} & \multicolumn{3}{c}{Rot. Error (${}^\circ$)} & \multicolumn{2}{c}{Mean} \\
\cmidrule(lr){3-5} \cmidrule(lr){6-8} \cmidrule(lr){9-10}
                        &                         & x      & y      & z      & roll      & pitch      & yaw      & Trans (m) & Rot (${}^\circ$) \\ \midrule
\multirow{3}{*}{eee\_01} & LI-Init                 & 0.018  & 0.406  & 0.074  & 0.235  & 1.635  & 3.464  & 0.166     & 1.778           \\
                        & L2Calib-g                & 0.064  & 0.058  & 0.072  & 0.738  & 1.661  & 1.715  & 0.064     & 1.371           \\
                        & L2Calib                    & 0.050  & 0.097  & 0.008  & 1.348  & 0.069  & 2.312  & \textbf{0.051} & \textbf{1.243}  \\ \midrule
\multirow{3}{*}{eee\_02} & LI-Init                 & 0.063  & 0.304  & 0.378  & 2.645  & 2.874  & 5.373  & 0.248     & 3.631           \\
                        & L2Calib-g                & 0.013  & 0.080  & 0.142  & 0.302  & 2.306  & 1.808  & 0.078     & 1.472           \\
                        & L2Calib                    & 0.044  & 0.049  & 0.090  & 1.204  & 0.208  & 2.134  & \textbf{0.061} & \textbf{1.362}  \\ \midrule
\multirow{3}{*}{eee\_03} & LI-Init                 & 0.426  & 0.082  & 0.193  & 8.730  & 19.751 & 15.709 & 0.234     & 14.730          \\
                        & L2Calib-g                & 0.140  & 0.091  & 0.114  & 3.058  & 0.794  & 1.480  & 0.115     & 1.777           \\
                        & L2Calib                    & 0.003  & 0.100  & 0.062  & 1.500  & 0.297  & 3.883  & \textbf{0.055} & \textbf{1.769}  \\ \midrule
\multirow{3}{*}{nya\_01} & LI-Init                 & 0.126  & 0.070  & 0.098  & 2.880  & 1.943  & 1.401  & 0.098     & 2.075           \\
                        & L2Calib-g                & 0.050  & 0.027  & 0.146  & 1.167  & 7.619  & 11.074 & 0.074     & 6.620           \\
                        & L2Calib                    & 0.030  & 0.045  & 0.016  & 2.839  & 1.536  & 1.013  & \textbf{0.030} & \textbf{1.796}  \\ \midrule
\multirow{3}{*}{nya\_02} & LI-Init                 & 0.578  & 0.139  & 0.290  & 0.528  & 2.518  & 0.798  & 0.336     & \textbf{1.281}  \\
                        & L2Calib-g                & 0.099  & 0.015  & 0.016  & 2.167  & 0.224  & 1.491  & 0.043     & 1.294           \\
                        & L2Calib                    & 0.047  & 0.039  & 0.003  & 3.162  & 0.062  & 3.204  & \textbf{0.030} & 2.143           \\ \midrule
\multirow{3}{*}{nya\_03} & LI-Init                 & 0.011  & 0.140  & 0.391  & 5.406  & 11.449 & 1.601  & 0.181     & 6.152           \\
                        & L2Calib-g                & 0.092  & 0.063  & 0.154  & 2.935  & 4.439  & 2.749  & 0.103     & 3.374           \\
                        & L2Calib                    & 0.031  & 0.083  & 0.017  & 0.222  & 4.642  & 1.306  & \textbf{0.044} & \textbf{2.056}  \\ \midrule
\multirow{3}{*}{sbs\_01} & LI-Init                 & ×      & ×      & ×      & ×      & ×      & ×      & ×         & ×               \\
                        & L2Calib-g                & 0.137  & 0.064  & 0.255  & ×      & ×      & ×      & 0.152     & ×               \\
                        & L2Calib                    & 0.109  & 0.020  & 0.124  & ×      & ×      & ×      & \textbf{0.084} & ×               \\ \midrule
\multirow{3}{*}{sbs\_02} & LI-Init                 & 0.314  & 0.213  & 0.060  & 6.963  & 53.790 & 27.587 & 0.196     & 29.447          \\
                        & L2Calib-g                & 0.048  & 0.093  & 0.100  & 3.256  & 0.715  & 24.692 & 0.080     & 9.554           \\
                        & L2Calib                    & 0.021  & 0.098  & 0.038  & 1.145  & 1.567  & 3.535  & \textbf{0.052} & \textbf{2.082}  \\ \midrule
\multirow{3}{*}{sbs\_03} & LI-Init                 & ×      & ×      & ×      & ×      & ×      & ×      & ×         & ×               \\
                        & L2Calib-g                & 0.037  & 0.028  & 0.012  & 1.526  & 0.585  & 15.676 & 0.026     & 5.929           \\
                        & L2Calib                    & 0.006  & 0.072  & 0.013  & 0.397  & 0.768  & 2.774  & \textbf{0.030} & \textbf{1.313}  \\ \midrule
\multirow{2}{*}{All Seq} & L2Calib-g                & 0.017  & 0.003  & 0.082  & 3.554  & 0.851  & 0.038  & 0.034     & 1.481           \\
                        & L2Calib                    & 0.040  & 0.003  & 0.035  & 1.362  & 0.183  & 0.013  & \textbf{0.026} & \textbf{0.519}  \\ \bottomrule
\end{tabular}
}
\end{table}

\subsection{UAV Dataset}

% For our experiments on UAV online flight data, we selected nine representative sequences from the NTU VIRAL dataset, divided into three distinct scenarios: eee with an average duration of 300 seconds, sbs with 371 seconds, and nya with 412 seconds. Each sequence exhibits low motion excitation, with average angular velocities below 0.15 rad/s. The eee and sbs sequences were recorded in outdoor environments, specifically a carpark and a square, while nya contains indoor data collected inside an auditorium. These long-duration sequences with weak excitation represent typical operational conditions for UAVs. We conducted extrinsic calibration between a horizontally mounted Ouster OS1-16 Gen 1 LiDAR and a VectorNav VN100 IMU using these online-collected datasets to evaluate the method's performance in real-world scenarios.

To evaluate our method under realistic UAV operational conditions, we conducted experiments using nine representative sequences from the NTU VIRAL dataset. These sequences cover three distinct scenarios: 
% eee (average duration: 300s), sbs (371s), and nya (412s). Each sequence exhibits weak motion excitation, with average angular velocities below 0.15 rad/s. 
the eee and sbs sequences were recorded in outdoor environments, specifically a carpark and a square, while nya contains indoor data collected inside an auditorium. We performed extrinsic calibration between a horizontally mounted Ouster OS1-16 LiDAR and a VectorNav VN100 IMU to assess the method’s robustness with weak excitation.

\begin{figure}[tb]
    \centering
    \includegraphics[width=\linewidth]{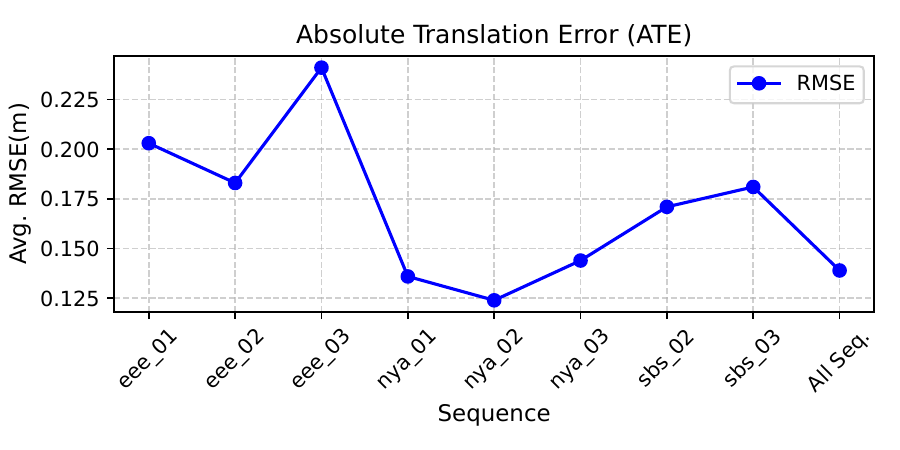}
    \caption{The RMSE between the estimated trajectory and the ground truth trajectory when running the odometry using calibration results from different sequences. The x-axis represents the sequence used for calibration, and the vertical axis represents the \textbf{average RMSE over ALL of the sequences}, sbs\_01 is not drawn. Note that the trajectory length for each sequence is approximately 2000 meters. Therefore, the obtained Absolute Translation Error, ranging between 0.12 and 0.23 meters, indicates that the odometry reliably operates.}
    \label{fig:rmse_ntu}
    \vspace{-1em}
\end{figure}

The corresponding calibration results are detailed in Tab. \ref{tab:ntucalib}. 
Despite the weak excitation, L2Calib achieves reliable 6DoF extrinsic calibration, albeit with a slight decrease in accuracy compared to the fully excited case. 
To further validate the usability of the calibrated extrinsic parameters, we applied them to the tightly-coupled odometry Fast-lio2 and the trajectory RMSE results are reported in Fig. \ref{fig:rmse_ntu}. The calibration accuracy is sufficient for odometry applications, ensuring stable trajectory estimation even under weak-excitation conditions.

% Furthermore, by \textbf{ individually applying the calibration results from each sequence to run odometry and testing trajectory accuracy across the entire dataset}, we demonstrate that the calibration results remain sufficiently reliable for odometry applications. The trajectory APE was expressed in Table \ref{tab:nturmse}. 

\begin{figure}[tb]
    \centering
    \includegraphics[width=\linewidth]{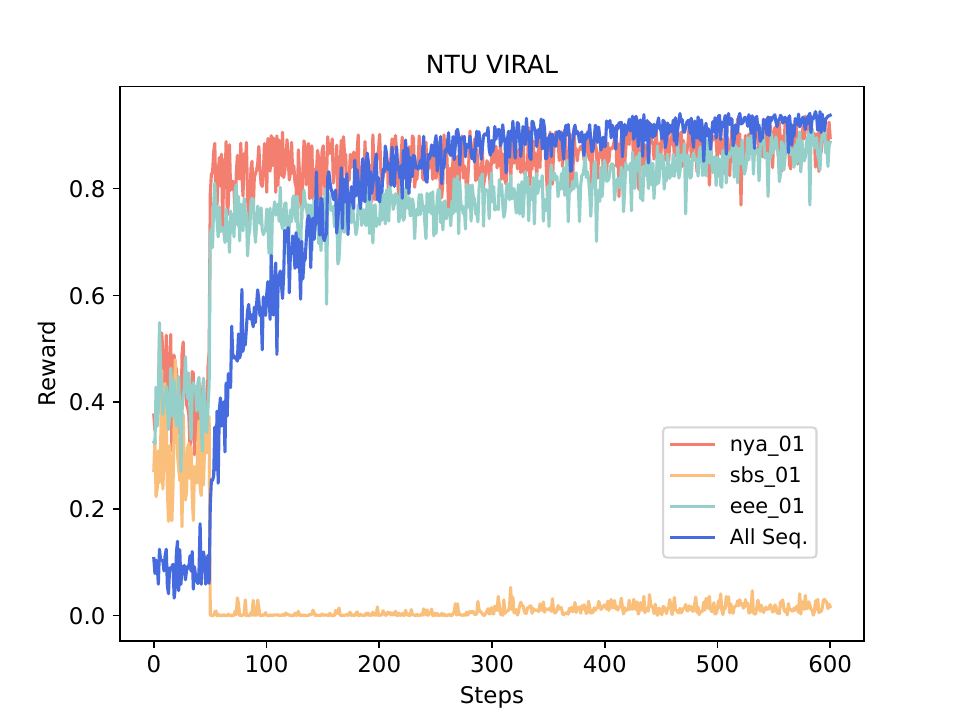}
    \caption{Comparison of convergence speed of different scenarios and using all sequences.}
    \label{fig:ntu_reward}
    \vspace{-1em}
\end{figure}

We analyze the impact of weak excitation on translation calibration. Due to insufficient excitation, single-sequence results show varying degradation across axes. As shown in Tab. \ref{tab:ntucalib}, eee\_01, eee\_03, and sbs\_02 exhibit significant y-axis errors (0.1m), consistent with the degradation trends in LI-Init. Indoor nya sequences yield lower errors, likely due to more reliable odometry in structured environments. In contrast, sbs\_01 fails in all methods due to overly simple motion, preventing robust extrinsic estimation. Multi-sequence calibration mitigates these issues by leveraging diverse motion patterns across different sequences, reducing the maximum translation calibration error to 0.04m and enhancing stability, improving odometry estimation over most single-sequence calibrations. The convergence analysis in Fig. \ref{fig:ntu_reward} reveals critical limitations of single-sequence calibration: in the sbs\_01 sequence, insufficient motion excitation leads to gradient vanishing that halts parameter updates, while sequences like nya\_01 and eee\_01 exhibit suboptimal convergence. In contrast, multi-sequence calibration achieves smoother convergence and superior optimization result by leveraging aggregated motion patterns.
% However, by compensating with data from other sequences, the overall usability of the calibration results is improved, the max error of translation is 0.04m. Moreover, the accuracy of the rotational extrinsic calibration is the highest.

\begin{table}[t]
\vspace{0.15cm}
\centering
\caption{Calibration errors on UGV dataset. Best results are in bold. ``All Seq'' represents that all sequences are used for calibration.}
\label{tab:mcdcalib}
\resizebox{\columnwidth}{!}{
\begin{tabular}{@{}c l c c c c c c c c@{}}
\toprule
\multirow{2}{*}{Seq}   & \multirow{2}{*}{Method} & \multicolumn{3}{c}{Trans. Error (m)} & \multicolumn{3}{c}{Rot. Error (${}^\circ$)} & \multicolumn{2}{c}{Ave. Error} \\
\cmidrule(lr){3-5} \cmidrule(lr){6-8} \cmidrule(lr){9-10}
                         &                         & x      & y      & z      & roll      & pitch      & yaw      & Trans (m) & Rot (${}^\circ$) \\ \midrule
\multirow{2}{*}{tu\_02} & LI-Init                 & 0.358  & 0.274  & 0.643  & 24.054 & 2.218  & 6.033  & 0.425     & 10.768          \\
                         & L2Calib                    & 0.044  & 0.081  & 0.006  & 1.737  & 0.160  & 0.117  & \textbf{0.044} & \textbf{0.671}  \\ \midrule
\multirow{2}{*}{tu\_03} & LI-Init                 & 0.119  & 0.767  & 0.795  & 8.775  & 3.211  & 1.287  & 0.560     & 4.424           \\
                         & L2Calib                    & 0.099  & 0.158  & 0.013  & 1.176  & 0.557  & 3.702  & \textbf{0.090} & \textbf{1.812}  \\ \midrule
\multirow{2}{*}{tu\_04} & LI-Init                 & 0.159  & 0.253  & 0.887  & 19.662 & 4.915  & 1.961  & 0.433     & 8.846           \\
                         & L2Calib                    & 0.083  & 0.019  & 0.003  & 0.177  & 0.178  & 2.660  & \textbf{0.018} & \textbf{1.005}  \\ \midrule
All Seq                  & L2Calib                 & 0.029  & 0.007  & 0.002  & 0.465  & 0.381  & 1.326  & \textbf{0.013} & 0.724           \\ \bottomrule
\end{tabular}
}
\end{table}
\begin{figure}[tb]
    \centering
    \includegraphics[width=\linewidth]{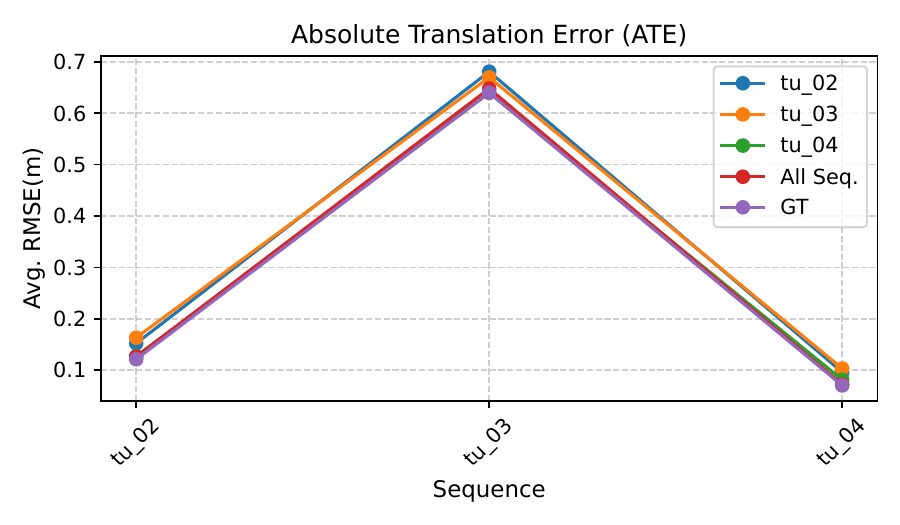}
    \caption{The RMSE between the estimated trajectory and the ground truth trajectory when running the odometry using calibration results from different sequences. The different colored lines indicate the results obtained using different extrinsic, and the horizontal axis indicates the sequence at which the odometry was performed.  GT represents that runs the odometry using ground truth calibration result.}
    \label{fig:rmse_mcd}
    \vspace{-1em}
\end{figure}

\subsection{Ground-Vehicle Dataset}

% 不失一般性，除了无人机数据集，我们还在无人车的数据集上进行了实验。在MCD数据集上选取了三个典型的序列进行实验，在数据集中，无人车运动的平均角速度为0.338弧度每秒 。三个序列的时长分别为500秒，839秒 and 187秒。我们所标定的是Ouster OS1 LiDAR和VN200 IMU之间的外参。我们仍采取对每个序列分别进行标定，然后用每个标定结果去运行里程计，在全部序列上计算轨迹APE的方式，来衡量外参标定的好坏。

% 从结果中我们可以得知，02中，y轴退化明显。03是更长的序列，且是动态环境，增大了标定难度，x、y轴的误差较大。但实际运行里程计的精度仍可用，虽然rmse达到0.31，但注意到真值的rmse达到0.28，所以影响不大。04中激励更全，使得误差更小。同样的，通过使用全部序列，整体标定可用性和全局一致性都提高了。

% Without loss of generality, in addition to the UAV dataset, we also conducted experiments on a UGV dataset. We selected three representative sequences from the MCD dataset for evaluation, where the average angular velocity of the UGV motion was 0.338 radians per second. The durations of the three sequences were 500 seconds, 839 seconds, and 187 seconds, respectively. The calibration focused on the extrinsic parameters between the Ouster OS1 LiDAR and the VN200 IMU. We adopted the same evaluation protocol as before: calibrating each sequence individually, running odometry with the calibrated parameters, and computing the Absolute Pose Error (APE) across all sequences to assess the quality of the extrinsic calibration.

To further validate our approach, we conducted experiments on a UGV dataset. We selected three representative sequences from the MCD dataset's tuhh\_day scenery (referred to as "tu" in this paper), calibrating extrinsics between Ouster OS1 LiDAR and the VN200 IMU.
The experimental results presented in Table \ref{tab:mcdcalib} demonstrate that the tu\_02 sequence exhibits significant degeneracy along the y-axis, likely attributable to insufficient motion excitation. For tu\_03, a longer sequence captured in dynamic environments, calibration proves more challenging, resulting in relatively larger errors in the x- and y-axis translations. Despite these challenges, the odometry performance remains viable for practical applications. To further validate the reliability of the calibration results, we executed odometry using the calibration results on each sequence. As illustrated in Fig. \ref{fig:rmse_mcd}, the RMSE between trajectories generated with our calibration results and those using ground-truth extrinsic remains minimal, confirming that the proposed calibration method maintains trustworthy odometry performance comparable to ground-truth configurations. 
By incorporating multiple sequences, calibration results become more stable, which aligns with the UAV experiments, demonstrating that our method effectively handles weak-excited data and remains applicable across various robotic platforms.

% From the results, we observe that in tu\_02, the y-axis exhibits significant degeneracy. The tu\_03 sequence is longer and collected in a dynamic environment, presents greater calibration challenges, resulting in larger errors along the x- and y-axes. Nevertheless, the odometry accuracy remains usable, with an RMSE of 0.31. Notably, the ground truth RMSE is 0.28, indicating that the impact of calibration errors is minimal. In tu\_04, more complete excitation leads to smaller errors. Similarly, by utilizing all sequences, the overall calibration usability and global consistency are improved. These findings align with the conclusions drawn from the UAV experiments, demonstrating that our method is applicable across various robotic platforms.

\begin{table}[t]
\vspace{0.15cm}
    \caption{Impact of data selection on calibration accuracy.}
    \label{tab:data_selection_ablation}
    \centering
        \begin{tabular}{l c c c}
            \toprule
            {Seq} & \makecell{{Data} \\ {Selection}}  & {Trans. Error} (m) & {Rot. Error} ($^\circ$) \\ 
            \midrule
            \multirow{2}{*}{eee\_01} & \ding{55} & 0.123 & 2.529 \\  
                                      & \ding{51} & \textbf{0.051} & \textbf{1.243} \\ 
            \midrule
            \multirow{2}{*}{eee\_02} & \ding{55} & 0.101 & 1.702 \\  
                                      & \ding{51} & \textbf{0.061} & \textbf{1.362} \\ 
            \midrule
            \multirow{2}{*}{eee\_03} & \ding{55} & 0.056 & \textbf{1.619} \\  
                                      & \ding{51} & \textbf{0.055} & 1.769 \\ 
            \bottomrule
        \end{tabular}
\end{table}

\begin{table}[t]
\vspace{0.3cm}
    \caption{Impact of different source of reference trajectory on calibration and tightly-coupled LIO estimation accuracy.}
    \label{tab:reftraj}
    \centering
        \begin{tabular}{c c c c c}
            \toprule
            % \textbf{Seq} & \textbf{Source} & \textbf{Trans. Error (m)} & \textbf{Rot. Error ($^\circ$)} & \textbf{RMSE} \\ 
            {Seq} & \makecell{{Trajectory} \\ {Source}} & \makecell{{Trans.} \\ {Error }(m)} & \makecell{{Rot.} \\ {Error}($^\circ$)} & {RMSE}(m) \\ 
            \midrule
            \multirow{2}{*}{eee\_01} & LO & 0.051 & 1.243 & 0.203 \\  
                                      & GT & \textbf{0.046} & \textbf{1.213} & \textbf{0.189} \\ 
            \midrule
            \multirow{2}{*}{eee\_02} & LO & \textbf{0.061} & 1.362 & 0.183 \\  
                                      & GT & 0.063 & \textbf{1.366} & \textbf{0.177} \\ 
            \midrule
            \multirow{2}{*}{eee\_03} & LO & 0.055 & 1.769 & 0.241 \\  
                                      & GT & \textbf{0.045} & \textbf{1.746} & \textbf{0.233} \\ 
            \bottomrule
        \end{tabular}
\vspace{-1em}
\end{table}

\subsection{Ablation study}

\subsubsection{\textbf{Policy Parameterization}}
% In this paper we discussed the unreasonableness of using Gaussian distribution to form the $S^3$ manifold, because the inherent bi-modality and symmetry of quaternions. To further express the effectiveness of Bingham distribution parameterization, we compare the Gaussian and Bingham sampling methods under different excitation levels and sequence types. The comparison calibration results accuracy were shown in Table \ref{tab:handholdcalib} and Table \ref{tab:ntucalib}, convergence speed were shown in Fig. \ref{fig:fully_reward}.

Using a Gaussian distribution for quaternion representation on the $S^3$ manifold is limited; we instead employ the Bingham distribution, which explicitly accounts for the symmetry of quaternions. Comparisons between L2Calib and L2Calib-g in Tab. \ref{tab:handholdcalib} and \ref{tab:ntucalib} show that Bingham-based sampling improves calibration accuracy across different motion excitations and scenarios, while Fig. \ref{fig:fully_reward} demonstrates its faster convergence. Unlike Gaussian distributions, where probability density is symmetric around the mean, quaternions require a representation where $q$ and $-q$ are equally probable, a property naturally modeled by the Bingham distribution. This leads to more stable calibration and improved convergence speed.

% In a Gaussian distribution, the probabilities of variables are symmetric around the mode, meaning that the probability of a value is the same on both sides of the mean. However, for quaternions, this property does not hold — the probability of a quaternion and its negative counterpart are equal. The Bingham distribution, on the other hand, is specifically designed to model this property, allowing it to better capture the intrinsic characteristics of rotation. This results in faster convergence and higher accuracy in calibration. 

\subsubsection{\textbf{Data Selection}}

We evaluate the impact of data selection on calibration accuracy and convergence speed, with results presented in Tab. \ref{tab:data_selection_ablation}. Selecting informative data significantly improves both accuracy and convergence speed, as freely recorded sequences often contain static or low-motion segments that are insensitive to extrinsic parameters. In such cases, even poorly estimated extrinsic parameters can result in a falsely high trajectory alignment reward, which misleads the reward calculation and degrades the calibration quality.

\subsubsection{\textbf{Reference Trajectory}}

The selection of reference trajectory introduces uncertainty in the calibration process. To analyze this effect, we conduct an ablation study comparing calibration accuracy and odometry performance when using two distinct reference trajectory sources: the LiDAR odometry (LO) trajectory and the ground truth (GT) trajectory from the Leica Prise system provided in the NTU VIRAL dataset. The results are summarized in Table \ref{tab:reftraj}. As anticipated, calibration utilizing the GT trajectory achieves superior calibration accuracy compared to the LO-based reference. However, odometry performance exhibits minimal discrepancies between the two approaches. This suggests that while GT references enhance calibration precision, both trajectory sources yield practically equivalent odometry reliability under the proposed framework.

\section{CONCLUSION}
In this paper, we highlight the absence of generic, user-friendly sensor calibration tools that can be easily applied across a wide range of sensor combinations. We propose a reinforcement learning-based framework to address this gap, allowing for extrinsic calibration with minimal data requirements and without the need for specialized calibration targets. Our method demonstrates strong performance in both fully-excited and weakly-excited conditions, significantly reducing the effort required for calibration. Experimental results show that our approach outperforms traditional methods in terms of calibration accuracy and flexibility, while also being adaptable to various sensor types and configurations. We believe our work paves the way for more efficient and autonomous calibration in robotic systems, and we plan to release our code to facilitate further research and development in this area.

\section{ACKNOWLEDGMENT}
This work was supported by the Key R\&D Project of Zhejiang Province under Grant 2025C01090. We would like to thank Kai Tang for helpful discussions and Yanming Wu for the assembly of data recording equipment. 

%\clearpage
{
% New Bib format
\AtNextBibliography{\scriptsize} % 
\printbibliography
}

\end{document}